\documentclass{article}

\usepackage{times}
\usepackage{graphicx} % more modern
\usepackage{subfigure} 

\usepackage{natbib}

\usepackage{algorithm}
\usepackage{algorithmic}

\usepackage{hyperref}

\usepackage[accepted]{icml2017}

\usepackage{graphicx}
\graphicspath{ {images/} }

\icmltitlerunning{Categorical Mixture Models on VGGNet activations (LE49 Michaelmas 2017)}

\begin{document} 

\twocolumn[
\icmltitle{Categorical Mixture Models on VGGNet activations (LE49 Michaelmas 2017)}



% you can specify symbols, otherwise they are numbered in order

\begin{icmlauthorlist}
\icmlauthor{Sean Billings}{cam}
\end{icmlauthorlist}

\icmlaffiliation{cam}{University of Cambridge, Cambridge, United Kingdom}

\icmlcorrespondingauthor{Sean Billings}{sb2219@cam.ac.uk}

\icmlkeywords{boring formatting information, machine learning, ICML}

\vskip 0.3in
]



\printAffiliationsAndNotice{}  % leave blank if no need to mention equal contribution

\begin{abstract} 

In this project, I use unsupervised learning techniques in order to cluster a set of yelp restaurant photos under meaningful topics. In order to do this, I extract layer activations from a pre-trained implementation of the popular VGGNet convolutional neural network. First, I explore using LDA with the activations of convolutional layers as features. Secondly, I explore using the object-recognition powers of VGGNet trained on ImageNet in order to extract meaningful objects from the photos, and then perform LDA to group the photos under topic-archetypes. I find that this second approach finds meaningful archetypes, which match the human intuition for photo topics such as restaurant, food, and drinks. Furthermore, these clusters align well and distinctly with the actual yelp photo labels. \\

\end{abstract}

\section{Introduction}

Image classification has received a lot of attention lately due to the emergence of successful deep convolutional neural network architectures. VGGNet, developed by \cite{vggnet}, along with its variants and outgrowths such as ResNet have performed exceptionally well on image classification competition tasks. In many cases, these algorithms achieve over 90\% accuracy, and sometimes with near perfect classification. It is strange that these techniques are so successful, but a delightful surprise. \\

A common image classification dataset is ImageNet. Models are trained to classify what object, such as 'hamburger', 'chair', or 'glass', is in a given photo. VGGNet proved to be exceptionally good at this task, and versions of VGGNet pre-trained on this dataset are freely available with the deep learning python library keras. \\

The ImageNet classification task demonstrates how neural networks like VGGNet have been largely successful for supervised learning tasks. However, we can test further how well  these systems understand images. In this paper, we imagine a scenario where we are given an unlabelled dataset, and we try to generate clusters of photos that roughly fall into meaningful 'topics'. We hope that unsupervised learning techniques can capture meaningful behavioural and content patterns in the kinds of photos that people take. In actuality, we have the class labels, but we force ourselves to perform unsupervised learning in order to see in which ways the intuition of the VGGNet neural network aligns with patterns in the ground truth. \\

In this paper, we examine the yelp restaurant photos dataset that is available freely online. We use the outputs/activations of various layers of VGGNet as features for Latent Dirichlet Allocation (LDA) models that will in turn produce photo 'topic' vectors and generate clusters of conceptually linked photos. \\

Section two of this paper will outline the theory behind convolutional neural networks. Section three of this paper will formalize a categorical mixture model that can be used for convolutional neural network activations and then more formally the latent dirchlet allocation model. Section four of this paper will outline and examine my experiments. Section five will conclude this paper and summarize this paper's findings. \\

\section{Training Convolutional Neural Networks} 

\subsection{Back Propagation}

Neural networks can be broken down into the combination of an input layer, a stack of hidden layers, and an output layer. The input layer connects an input vector x to our first hidden layer along with a bias term. Each hidden layer at depth k of our network is parameterized by a set of weights $ w^k_{ij} $ defining a set of activations $a^k_j = \sum w_{ij}z^k_i $ and a set of nonlinear functions $g^h(a_{j}) $ that act on those activations in order to generate the input $z^{k+1}_i$ to the next layer of our network. For regression problems, the output layer of our neural network computes a single real value $ \hat{y} $ that minimizes a loss function $ L(\hat{y},t) $ which measures the distance from the output value of a network to a target value $t$. For classification problems, we often define a loss function such as cross-entropy in order to ensure that the greatest number of training examples are classified correctly. \\

If we explicitly define our loss function we can then compute an error $E_n = L(y(x_n),t_n))$ for each data point $x_n$ in our data set, where $y(x_n)$ is our network output and $t_n$ is our target from the training dataset. The backpropagation algorithm (see Algorithm 1) works by performing a forward pass over the network to compute the errors $ E_n $, and then a backwards pass over the network in order to compute the derivatives $ \frac{\partial E_n}{\partial w^k_{ij}}$ for each layer of the network. For a convolution neural network, we must enforce weight constraints at the convolutional layers of our network. To do this, we can augment for the tied weights according to our convolutions. To do this we can augment our calculated error derivatives with $ \frac{\partial E_n}{\partial w_a} = \frac{\partial E_n}{\partial w_b} = \frac{\partial E_n}{\partial w_a} + \frac{\partial E_n}{\partial w_b}$. This way, all tied weights $w_a$, $w_b$ will be updated equally throughout the back-propagation optimization. \\

Formally, we can break down the training of a Convolutional Neural Network (CNN) such as VGGNet \cite{vggnet} with back-propagation as follows: For a classification problem, we define the error function for the log likelihood for our convolutional neural network with the cross entropy error function \cite{bishopnn} 

\begin{equation}
E_{CNN} = - \sum (  t^n \ln y^n + ( 1-t^n ) \ln (1 - y_n)
\label{eq:errordae}
\end{equation}

Using the chain rule, we can derive the form of the derivative of the error function with respect to the weights as 

\begin{equation}
\frac{\partial E_{CNN}}{\partial w^k_{ij}}=  \frac{\partial E_{CNN}}{\partial a^k_{j}} \frac{\partial a^k_{j}}{\partial w^k_{ij}} 
\label{eq:ederivativedae}
\end{equation}

where $ \frac{\partial a^k_{j}}{\partial w^k_{ij}} = z^k_i$. As in \cite{bishopnn}, we use the notation $\delta^k_j = \frac{\partial E_{CNN}}{\partial a^k_{j}} $. This is used for convenience to represent the error differences at layer $k$. This allows us to explicitly represent and compute the error difference computations at each layer of our neural network.  \\

For the output layer, we compute the error differences as

\begin{equation}
\delta^k_j = \frac{\partial E_{CNN}}{\partial y_n} = \frac{y_n-t_N}{y_n(1-t_n)}
\label{eq:outputdj}
\end{equation}

For the hidden layers, we compute the error differences as

\begin{equation}
\delta^k_j = g^{k'}(a_j) \sum_i w^{k+1}_{ij} \delta^{k+1}_i
\label{eq:outputdj}
\end{equation}

where we iterate over the nodes $i$ in the $k+1$ layer that the node $j$ in the kth layer is connected to.\\

\begin{algorithm}[tb]
   \caption{Backpropagation for Convolutional Neural Network}
   \label{alg:example} 
\begin{algorithmic}
   \STATE {\bfseries Input:} Network \{network structure g , initial weights w\}, learning rate $\eta$, $\{x_n\}$ training examples
   \STATE {\bfseries Output:} Network \{weights $w^*$\}
   \FOR{ $iterations$}
   \FOR{ $x_n$ in training examples}
   \STATE Forward pass to compute $ E_n $
   \STATE Backward pass to compute $\nabla E_n = \frac{\partial E}{\partial w^k_{ij}} $
   \FOR {Tied weights $w_a$ , $w_b$ }
   \STATE set $ \frac{\partial E_n}{\partial w_a} = \frac{\partial E_n}{\partial w_b} = \frac{\partial E_n}{\partial w_a} + \frac{\partial E_n}{\partial w_b}$
   \ENDFOR
   \STATE update $ w^k_{ij} = w^k_{ij} - \eta \nabla E_n$
   \ENDFOR
   \ENDFOR
\end{algorithmic}
\end{algorithm}

As shown above, the gradient of one layer depends solely and simply on the activation functions and the error of the next layer. These derivatives are fed into a gradient based optimization algorithm in order to train the neural network. These gradients can be computed efficiently, especially when taking into consideration GPU hardware. Taking a naive approach, the number of iterations required to train a strong network can still be very high, and a lot of compute power can be required. However, if we efficiently store our convolutional layers, we can actually improve our performance significantly. \\

\subsection{Convolutional Layers}

\begin{table*}[t]
\caption {VGGNet-16 Architecture}
\label{tab:vggnet}
\begin{center}
\begin{tabular}{ |c|c|c|c|c|c|c|c|c|c|c|} 
\hline
input & cov3-64 & conv3-128 & conv3-256 & conv3-512 & conv3-512 & dense-4096  \\
& cov3-64 & conv3-128 & conv3-256 & conv3-512 & conv3-512 & dense-4096 \\
& maxpool & maxpool    & conv3-256 & conv3-512 & conv3-512 & dence-1000 \\
& 		 &  		     & maxpool & maxpool & maxpool & softmax \\
\hline
\end{tabular}
\end{center}
\end{table*}

Convolutional Network layers are a form of weighted tensor-based feature maps developed by \cite{lecuncnn}. Recently, they have seen prominence in many competition winning image classification systems including VGGNet \cite{vggnet}. Formally, a convolutional network layer defines a convolution (or sometimes a cross-correlation) which is essentially a mapping from an input layer and a defined kernel into the output layer. As in \cite{deeplearningbook}, for a two dimensional image $I$ and kernel $K$, we can define the convolution $S(i,j) $ as follows, 

\begin{equation}
S(i,j) = (K*I)(i,j) = \sum_m \sum_n I(i-m,j-n) K(m,n)
\label{eq:convolution}
\end{equation}

The convolution $S(i,j) $ is an activation taken over the $m$ and $n$ neighbouring pixels to the $(i,j)$ pixel in the image I. We train the $m$ x $n$ weights of the kernel $K(m,n)$ using backpropagation in order to generalize to the entire layer input. In other words, the kernel $K$ that is trained is identical for every connection at that layer of the network. This is achieved using tied-weights. Using a convolutional layer as an alternative to a fully connected layer is useful because it greatly reduces the number of required training parameters. This reduces the storage space required and allows us to architect much deeper networks. Furthermore, sequences of convolutions can act as meaningful collections of filters that are particularly useful for finding abstract textures and shapes in images.  \\

\subsection{Pooling}

A pooling layer in a convolutional neural network enforces that the representation of the data at that layer is invariant to small translations of the input data. A standard pooling metric is to take the maximum activation from a set of nearby activations. This is particularly useful for enforcing the neural network to perform feature detection that is not overly dependent on the location of a given feature \cite{deeplearningbook}. Pooling layers are usually included in a network after convolutional layers. The VGGNet architecture makes use of such pooling strategies after all of its sets of convolutional layers \cite{vggnet}. Max-pooling layers are architecturally nice because they introduce no additional weights to the network. \\

\subsection{VGGNet}

The success of VGGNet, developed by the Oxford Visual Geometry Group \cite{vggnet}, is part of the start of a transition from shallow wide neural networks to deep network architectures. It is still not exactly known formally why deep layers seem to be more effective than shallow neural networks. One can even show that two-layer fully connected neural networks are capable of representing any function \cite{bishopnn}. However, the recent most successful competition architectures have been deep. Deep networks seem to train more effectively. Therefore, in the common perspective, deep networks have become favoured over thick shallow networks. \\ 

The core units of the VGGNet architectures are pairs or triplets of 3-unit Convolution layers followed by max pooling layers. This ensures a stable translation invariance as the network produces features. The final layers of the VGGNet architecture are three fully connected layers mapping in to a final Softmax activation. This paper makes use of the 16 layer VGGNet-16 architecture. The full architecture can be seen in table \ref{tab:vggnet}. \\

The implementation of VGGNet used in this paper has been pre-trained on the ImageNet dataset. The ImageNet dataset is part of a competition photo classification task that looks to predict which object out of 1000 labels a photo actually consists of. This is actually somewhat miss-aligned with the yelp photo dataset classification. The yelp photo dataset consists of 5 potential labels 'food', 'inside', 'menu', 'drink', and 'outside. These labels cover and relate to broad sets of objects in the ImageNet dataset. However, by treating the imageNet outputs and activations as topic components for sufficiently trained categorical models, we can discover a latent topic representation of ImageNet objects that align well with the Yelp dataset labels. \\

\section{Categorical Mixture Models of Convolutional Features}

Categorical mixture models can be used to discover the latent structure of a dataset. For the yelp photos, we assume that the images in our dataset are each related to one of $t$ topics, and that each of these topics is associated with its own categorical distribution over the images in that topic. We parameterize our model such that we can estimate the probability that the activations for each image correspond to the distribution for a given topic. \\

\subsection{Maximum Likelihood Estimate for the Categorical Mixture Model} 

Formally,  let us represent the $j$ convolutional activations of the k'th layer of our neural network acting on the i'th image as $S^{(k)}_j(i) $. We require two sets of parameters for our categorical mixture model. \\

First, we require a set of $t$ vectors $ [ \beta_0 ... \beta_t ] $ that parameterize our categorical model for the probability of observing (for a fixed layer k) the activation $S^{(k)}_j(i) $ given that the image $ i$ was assigned to topic $t$. This conditional probability is expressed as $ p(S^{(k)}_j(i)  | z_i = t) $ and is categorically distributed over $\beta_{t}$ according to

\begin{equation}
 p(S^{(k)}_j(i)  | z_i = t, \beta_t) = \prod_j \beta_{tj}^{S^{(k)}_j(i)} 
\label{eq:catbeta}
\end{equation}

Secondly, we require a representation of the probability of assigning image $i$ to topic $t$. We parameterize this probability with the vector $\theta$ such that, $ p(z_i = t) = \theta_t $. The likelihood for the data given our model $p(S^{(k)} | \theta,\beta )$ over images $i$, topics $t$, and activations $j$ for the k'th layer of our network is then

\begin{equation}
p(S^{(k)} | \theta,\beta ) = \prod_i \sum_t \theta_t \prod_j \beta_{tj}^{S^{(k)}_j(i)} 
\label{eq:catlikelihood}
\end{equation}

\subsection{Latent Dirchlet Allocation} 

By assigning priors to the parameters $\theta$, $\beta$ of our model, and by changing our assumptions about how documents are generated under our model, we can evolve our categorical mixture model into a formulation of the well known Latent Dirichlet Allocation (LDA) approach. Formally, we attach Dirichlet priors on $\theta$ and $\beta$, where 
$\theta$ is parametized by $\alpha$ according to the dirichlet prior $\theta = \frac{1}{B(\alpha)} \prod_t z_t^{\alpha_t} $ and $\beta_t$ for topic $t$ is parametized by $\gamma $ over actings $S_j$ such that that $\beta = \frac{1}{B(\gamma)} \prod_j S_j(i) ^{\gamma_j} $. For the dirichlet distributions above $B(...)$ is the normalizing Beta distribution. The other change to our model is to augment the document-topic assignment $z_t$ to a activation-topic assignment $z_{jt}$. In this model each activation $S_j$ for a given image $i$ is said to be generated according to a single topic $t$, and the image overall as a mixture of topics. \\

\section{Experiments}

The goal of this section is to evaluate several image clustering techniques and speculate about the resulting distributions generated. We examine two neural network and probabilistic model combinations. The neural network is an out-of-the-box VGGNet implementation from Keras trained on the ImageNet dataset. We run our probabilistic models on the freely available Yelp photos dataset. The main purpose of these experiments is to demonstrate the power of \textit{unsupervised} learning techniques. To do this, we look at how closely our unsupervised clusters follow the distribution for unobserved photo labels. \\

\subsection{Latent Dirchlet Allocation on VGGNet Convolutional Activations}

Firstly, we pass our yelp photos dataset through VGGNet in order to extract top-level convolutional activations for each image. We then run LDA for k topics on the top level activations, treating each activation node as a separate word. We set an activation threshold in order to regulate the number of activations that are considered 'on'. In practice this has a great impact on performance. Using a threshold of $S^k_{ij} > 0$ will tend to create vague LDA topic archetypes, whereas higher activation thresholds tend to produce more distinct topics. The hyper-parameters for this experiment are the activation threshold and the activation layer chosen to use as features. One can also consider further training of the network with the annotated yelp photos, although I do not present that here. \\

Below, I show the label concentrations for $k=4$ LDA topics generated from a sample of approximately 16000 yelp photos using ImageNet trained VGGNet convolutions with an activation threshold of 100. I found that thresholds of 50, or 10 produced more vague topics, and an activation threshold of 0 produced unsatisfactory topics with even the top features for each topic having weights very close to 0. \\

\begin{table}[h]
\caption {Density of image labels for CNN activation LDA topics}
\label{tab:LDAConvTopics}
\begin{center}
\begin{tabular}{ |c|c|c|c|c|c|c|c|c|c|c|} 
\hline
topics & food & menu & inside & drink & outside \\
 \hline
0 (food) & 5095 & 1 & 40 & 2 & 9 \\
\hline
1 (drink+ $\epsilon$)  & 1703 & 6 & 328 & 552 & 108 \\
\hline
2 (food 2) & 3122 & 45 & 137 & 1 & 17 \\
\hline
3 (restaurant) & 90 & 34 & 3545 & 11 & 1538 \\
\hline
\end{tabular}
\end{center}
\end{table}

It is very interesting that we can build topics from convolutional activations that correlate with the photo labels without observing the photo labels directly. Training on the Yelp photos is not even necessary. One setback for understanding this approach is that the topic vectors are composed of activation labels that are not exactly meaningful. For example, it is difficult to figure out in which ways exactly topics 0 and topics 2 differ without inferring from a sample of pictures. We address this deficiency in the next section by looking at predicted objects as features. \\ 

\subsection{Latent Dirchlet Allocation Clustering on ImageNet Trained Image Labels}

For our second experiment, we combine the classification power of VGGNet trained on ImageNet with unsupervised learning from LDA in order to generate feature-topic archetypes. We use the top 10 output activations of VGGNet for object detection in the image. VGGNet is able to pick out objects such as 'glass', 'hotdog', and 'restaurant' that end up being meaningful descriptions of the objects in a given image. After collecting these object predictions as features of each image, I run LDA in order to cluster each image into one of 5 generated topic archetypes. We use one additional topic in this experiment compared to the last for experimentation purposes.  I find that the generated topics revolve around the different kind of identifiable shots in a restaurant, and that there is a more intuitive divide between the various topics than in the previous experiment.  \\ 

\begin{table}[h]
\caption {Density of image labels for ImageNet object LDA topics}
\label{tab:LDAObjectTopics}
\begin{center}
\begin{tabular}{ |c|c|c|c|c|c|c|c|c|c|c|} 
\hline
topics & food & menu & inside & drink & outside \\
 \hline
0 (restauraunt) & 33 & 3 & 3173 & 4 & 1089 \\
\hline
1 (food)  & 1345 & 1 & 67 & 40 & 8 \\
\hline
2 (food 2) & 3582 & 17 & 108 & 158 & 5 \\
\hline
3 (food 3) & 4395 & 0 & 12 & 5 & 3 \\
\hline
4 (drink+menu) & 655 & 65 & 690 & 359 & 567 \\
\hline
\end{tabular}
\end{center}
\end{table}

\begin{figure}[t]
\begin{minipage}{0.24\textwidth}
    \includegraphics[width=\linewidth]{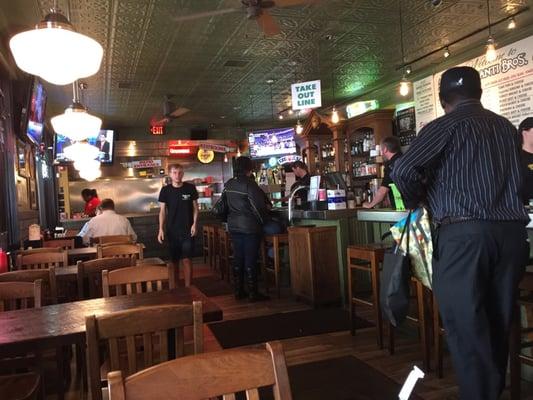}
    \end{minipage}
    \hspace{\fill} % note: no blank line here
    \begin{minipage}{0.22\textwidth}
    \includegraphics[width=\linewidth]{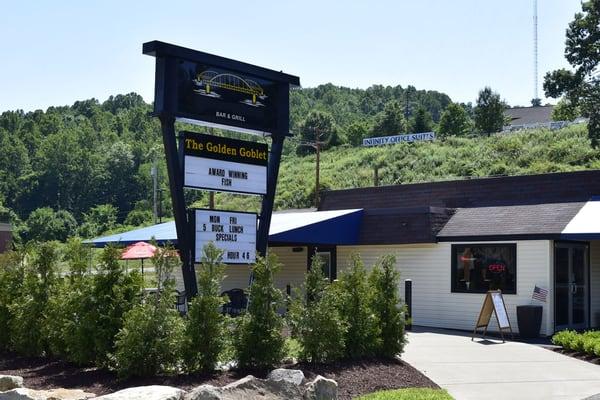}
    \end{minipage}

    \vspace*{1cm} % vertical separation

    \begin{minipage}{0.25\textwidth}
    \includegraphics[width=\linewidth]{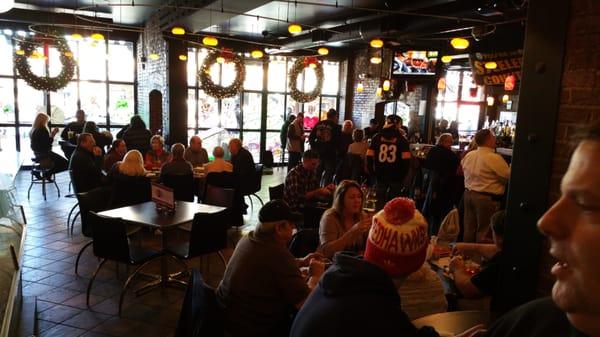}
    \end{minipage}
    \hspace{\fill} % note: no blank line here
    \begin{minipage}{0.2\textwidth}
    \includegraphics[width=\linewidth]{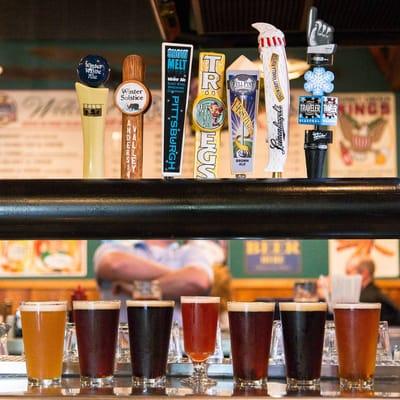}
    \end{minipage}
\caption{Images from Restaurant Topic} 
\label{fig:4restaurant}
\end{figure}

In Figure \ref{fig:4restaurant}, I plot four chosen sample images from topic 0 from experiment 1. These photos are pretty clearly restaurant photos, although the fourth photo might ambiguously be determined as drink, although it is labeled as 'inside' in the yelp dataset. The density for restaurant images of this cluster is very high, so it is very likely that a random photo chosen from this cluster will be a restaurant photo. 

Because the topic archetypes are composed of meaningful features such as 'hamburger', 'hotdog', 'guacamole', or 'pizza', we can more easily discern what kind of photos actually belong to each cluster. In table \ref{tab:LDATopicObjects}, I display some of the most influential vector components for each topic. This also allows us to clarify some of the labels in table \ref{tab:LDAObjectTopics}. For example, we can see that topic 2 is actually composed of mostly desert terms, cluster 3 is mostly composed of full meal terms, and topic 1 is mostly soup related terms. This available clarity is markedly different from the first experiment since the object labels are so much more intuitively meaningful.  \\

\begin{figure}[h]
\begin{minipage}{0.2\textwidth}
    \includegraphics[width=\linewidth]{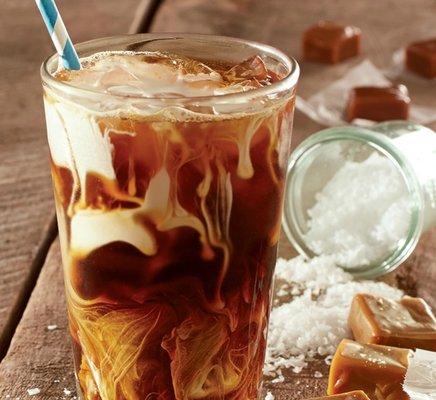}
    \end{minipage}
    \hspace{\fill} % note: no blank line here
    \begin{minipage}{0.2\textwidth}
    \includegraphics[width=\linewidth]{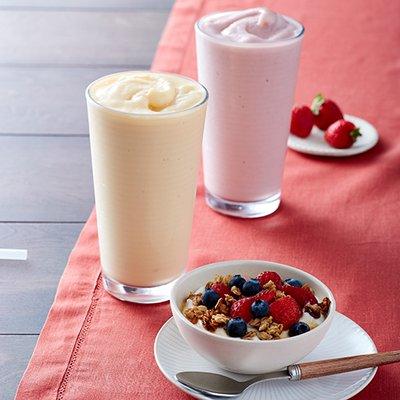}
    \end{minipage}

    \vspace*{1cm} % vertical separation

    \begin{minipage}{0.22\textwidth}
    \includegraphics[width=\linewidth]{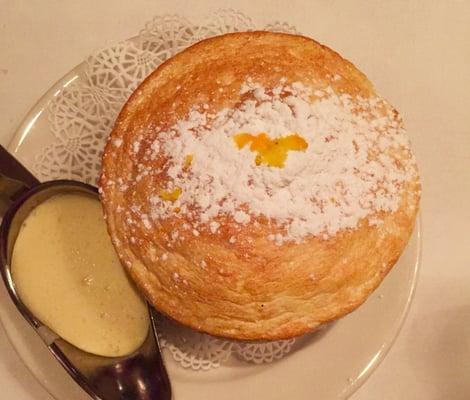}
    \end{minipage}
    \hspace{\fill} % note: no blank line here
    \begin{minipage}{0.2\textwidth}
    \includegraphics[width=\linewidth]{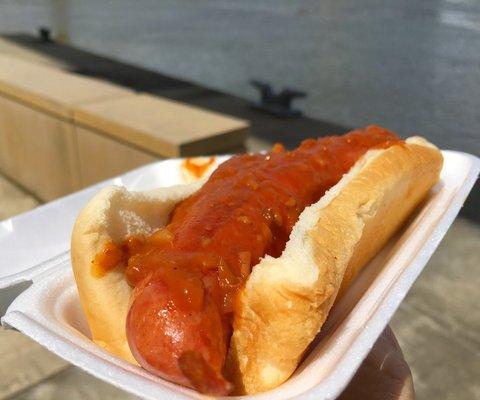}
    \end{minipage}
\caption{Images from Dessert Topic} \label{fig:4drinks}
\end{figure}

For this experiment, topics 0, 1, and 3 seem pretty well clustered in terms of distinct yelp photo labels. However, topics 2 and 4 are notably noisy. At first glance, we can consider this a deficiency of our algorithm, but there may be something more going on here. If we look at a sample of photos in cluster 2 (shown in figure \ref{fig:4drinks}), we find that the drink-labelled photos in this cluster are actually very dessert-like. It seems that our LDA model has picked up on some meaning that would otherwise not be available in the class labels. Actually, we may be misinformed by the class labels to not recognize those drinks as dessert-like items. Despite this interesting insight, the cluster is still badly noisy. There are many other types of non-food object labels, and, strangely, 'hotdog' is actually one of the top ten features for this topic. This topic is a good example for both for how the unsupervised LDA can find some hidden meaning, and also that it might make topic decisions that we would see as strange and unnatural. \\

\begin{table*}[t]
\caption {LDA topic sample components}
\label{tab:LDATopicObjects}
\begin{center}
\begin{tabular}{ |c|c|c|c|c|c|c|c|c|c|c|c|} 
\hline
topics & good & good  &  good & good  &  good & bad\\
 \hline
0 (restaurant) & restaurant & bakery & dining table & barbershop & bookshop & confectionery \\
\hline
1 (food in bowls) & hotpot & soup bowl & wooden spoon & ladle & mortar & spatula \\
\hline
2 (dessert) & plate & ice cream & chocolate sauce & trifle & tray  & hotdog \\
\hline
3 (meals) & plate & guacamole & burrito & hotdog & carbonara & \\
\hline
4 (drinks) & wine bottle & beer glass & goblet & cocktail shaker & beer bottle & palace\\
\hline
\end{tabular}
\end{center}
\end{table*}

The components in table \ref{tab:LDATopicObjects} are a sample of the top components for each LDA topic. We can see that running LDA with ImageNet features has produced mostly meaningful topics, although some bad features in the topics are highlighted. \\

The kinds of distributions found in the two experiments are vaguely similar, but certain class labels are handled differently. For example, we can see that the menu category is clustered much more purely in the second experiment than the first, while drinks are clustered much more uniformly in the first approach.  There are also some similarities. For example, both examples have well partitioned restaurant topics, as well as at least one well partitioned food topic. It is interesting to note that both approach had difficulty in their purifying their respective 'drink' topics. The LDA component of our probabilistic model seems to pick up on archetypes for the types of photos people like and are likely to take. This learned pattern is the actual human photo choice behaviour rather than the annotated class labels. \\

Considering that the LDA module is run on entirely different features for the two experiments, the types of clusters produce by the two experiments are actually pretty similar. This would make sense if both the convolutional activations and the output labels are similarly strongly correlated with these latent features. This is not entirely unreasonable, as both the output and convolutions belong to the same network stack, and are closely co-located. On the other hand, there are several fully connected network layers in between, so it is not obvious that the latent representation would be so similar. This can be seen as evidence that the LDA component is picking up some meaningful latent features of the photo distribution. 

\section{Conclusion}

In this paper, I showed how the activations of a pre-trained neural network can be used as a base for unsupervised learning on a novel space. Furthermore, I provided evidence that the cluster topics generated by this approach are meaningful and can accurately pick up on an underlying distribution of class labels. We looked at two ways of using neural network activations as features for LDA. First, we considered using directly the final convolutional activations of the network. Secondly. we considered feeding the top VGGNet outputs as features into LDA. I found that both strategies produce similar distributions, and infer that both strategies might be looking at a similar underlying latent structure. I also find that using the VGGNet outputs is a more descriptive and intuitive methodology. \\

Performance is difficult to measure in this scenario, because in a sense we are trying to discover an interpretation of the data that is more rich than the provided photo labels. For example, it seems that the LDA topics find more meaningful distinctions under the food labeled photos, and the inside-outside distinction was less useful to LDA . I think therefore, the real way to improve performance (in terms of understanding, rather than class label prediction) is to train a model that has a more deep understanding of the different objects that appear in the yelp photos. Potential ways to improve this understanding could include using more recently developed CNN architectures such as ResNet and cross-validation of priors for LDA. \\

We have validated that the unsupervised learning in this paper provides a fairly successful approximation of learning the ground truth class labels. however,  I do think that the more fruitful line of effort is to develop better image understanding. A very interesting offshoot of this type of system would be a generative solution to this problem. The linear topics of LDA are useful for clustering, but what would be more generally useful would be a more sophisticated non-linear and generative topic model. For this generative case, we would want to be able to develop topic photos that reflect the types of objects that would be present in the topic vectors learned here. \\

Class labels usually make data simple to deal with because the training task becomes well defined. Supervised learning has become almost a synonymous anecdote for machine learning, and it makes it easy forget that we do not always have class labels and meaningful annotations. Evaluating and understanding human photo patterns through unsupervised learning is generally not as defined or concise a task as photo classification. Therefore, naturally, a lot of what I infer in this paper is speculative. It is an attempt to manage vague notions and strange results. This paper can be seen as reflection and surprise for the powerful generalizability of neural network architectures, even for tasks that the systems were not  designed for. \\

\bibliography{Categorical_Mixture_Models_on_VGGNet_Conv_Features}

\begin{thebibliography}{4}
\providecommand{\natexlab}[1]{#1}
\providecommand{\url}[1]{\texttt{#1}}
\expandafter\ifx\csname urlstyle\endcsname\relax
  \providecommand{\doi}[1]{doi: #1}\else
  \providecommand{\doi}{doi: \begingroup \urlstyle{rm}\Url}\fi

\bibitem[Bishop(1995)]{bishopnn}
Bishop, Christopher~M.
\newblock \emph{Neural Networks for Pattern Recognition}.
\newblock Oxford University Press, Inc., New York, NY, USA, 1995.
\newblock ISBN 0198538642.

\bibitem[Goodfellow et~al.(2016)Goodfellow, Bengio, and
  Courville]{deeplearningbook}
Goodfellow, Ian, Bengio, Yoshua, and Courville, Aaron.
\newblock \emph{Deep Learning}.
\newblock MIT Press, 2016.
\newblock \url{http://www.deeplearningbook.org}.

\bibitem[LeCun \& Bengio(1998)LeCun and Bengio]{lecuncnn}
LeCun, Yann and Bengio, Yoshua.
\newblock The handbook of brain theory and neural networks.
\newblock chapter Convolutional Networks for Images, Speech, and Time Series,
  pp.\  255--258. MIT Press, Cambridge, MA, USA, 1998.
\newblock ISBN 0-262-51102-9.
\newblock URL \url{http://dl.acm.org/citation.cfm?id=303568.303704}.

\bibitem[Simonyan \& Zisserman(2014)Simonyan and Zisserman]{vggnet}
Simonyan, Karen and Zisserman, Andrew.
\newblock Very deep convolutional networks for large-scale image recognition.
\newblock \emph{CoRR}, abs/1409.1556, 2014.
\newblock URL \url{http://arxiv.org/abs/1409.1556}.

\end{thebibliography}
\bibliographystyle{icml2017}

\end{document}